\documentclass[runningheads]{llncs}

\usepackage{accv}
\usepackage{marvosym}

\usepackage{accvabbrv}

\usepackage{graphicx}
\usepackage{booktabs}
\usepackage{multirow}
\usepackage[table]{xcolor}
\usepackage{pifont}
\usepackage{tcolorbox}
\tcbuselibrary{skins,breakable}
\usepackage{algorithm}
\usepackage{algorithmic}
\usepackage[accsupp]{axessibility}  

\newcommand{\rev}[1]{#1}

\usepackage{hyperref}

\begin{document}

\title{Does Peer Observation Help? Vision-Sharing Collaboration for Vision-Language Navigation} 

\titlerunning{Co-VLN}

\author{Qunchao Jin\inst{1} \and
Yiliao Song\inst{1} \and
Qi Wu\inst{1,2}\textsuperscript{(\Letter)}}

\authorrunning{Q.~Jin et al.}

\institute{Australian Institute for Machine Learning, Adelaide University, Adelaide, Australia \and
Responsible AI Research Centre, Australian Institute for Machine Learning, Adelaide, Australia\\
\email{\{qunchao.jin, lia.song, qi.wu01\}@adelaide.edu.au}}

\maketitle

\begingroup
\renewcommand{\thefootnote}{}
\footnotetext{\Letter\ Corresponding author. Project page: \url{https://QunchaoJin.github.io/CoVLN/}}
\endgroup

\begin{abstract}

Vision-Language Navigation (VLN) systems are fundamentally constrained by partial observability, as an agent can only accumulate knowledge from locations it has personally visited. As multiple robots increasingly coexist in shared environments, a natural question arises: can agents navigating the same space benefit from each other's observations? In this work, we introduce Co-VLN, a minimalist framework for systematically investigating whether and how peer observations from concurrently navigating agents can benefit VLN. When independently navigating agents identify common traversed locations, they exchange structured perceptual memory, effectively expanding each agent's receptive field without any additional exploration of its own. We validate our framework on the R2R benchmark under two representative paradigms (the learning-based DUET~\cite{chen2022think} and the zero-shot MapGPT~\cite{chen2024mapgpt}), \rev{and further show that it generalizes to continuous environments with ETPNav~\cite{an2024etpnav}}. We also conduct extensive analytical experiments to systematically reveal the underlying dynamics of peer observation sharing in VLN. Results demonstrate that vision-sharing enabled model yields substantial performance improvements across both paradigms, establishing a strong foundation for future research in collaborative embodied navigation.

\end{abstract}

\section{Introduction}
\label{sec:intro}

Over the past few years, Vision-Language Navigation (VLN) has witnessed remarkable progress, largely driven by increasingly sophisticated architectures, ranging from RNN-based seq2seq models~\cite{anderson2018vision,fried2018speaker,wang2019reinforced}, to pretrained Vision–Language Transformers~\cite{majumdar2020improving,chen2022think,hong2021vln,chen2021history,wang2023scaling,qiao2023hop+}, and more recently, to foundation models~\cite{cheng2024navila,wei2025streamvln,zeng2025janusvln,zhang2025embodied} that unify perception, grounding, and planning. However, these advancements primarily focus on architectural designs, inherently defaulting to egocentric observations for guiding navigation. Consequently, VLN systems remain constrained by partial observability, the agent must make decisions with incomplete knowledge of the environment, which bottlenecks performance particularly in complex and long-horizon tasks.

\begin{figure*}[t]
    \centering
    \includegraphics[width=0.9\textwidth]{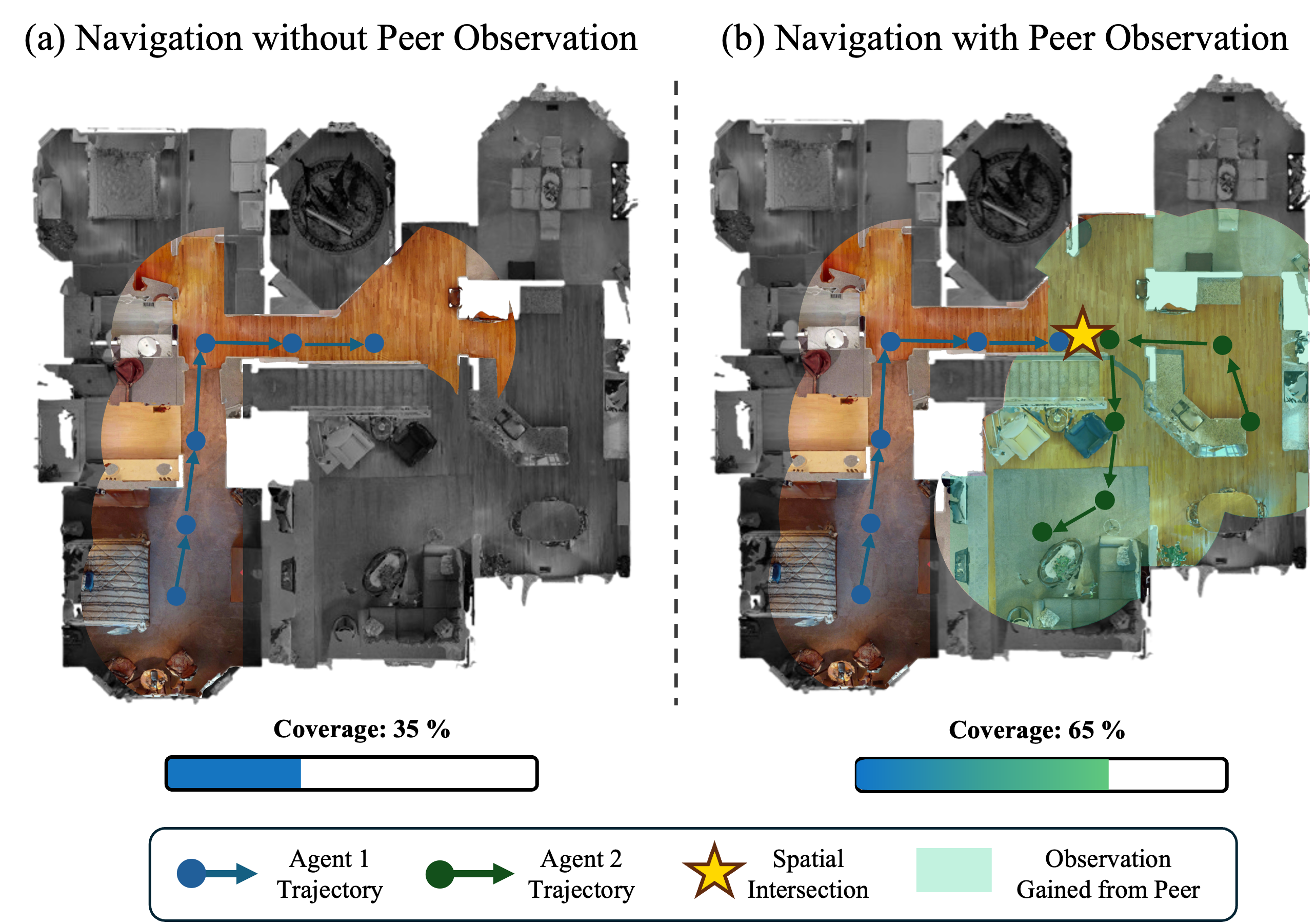}
    \caption{\textbf{Motivation of peer observation sharing in VLN.} (a) A single agent navigates independently, with its perception limited to personally visited regions while large portions of the environment remain unobserved, resulting in only 35\% spatial coverage. (b) When a peer agent independently navigates the same environment, their trajectories may overlap at a spatial section (yellow star), enabling knowledge exchange that expands perceptual coverage from 35\% to 65\% without additional exploration.}
    \label{fig:intro}
\end{figure*}

A natural response to this challenge has been to enrich the representation of the environment. Prior works have pursued this through various strategies, such as backtracking mechanisms that allow revisiting previously observed regions~\cite{ke2019tactical, ma2019regretful}, structured memory construction via topological graphs~\cite{an2023bevbert,chen2022think},  and predicting imagination of unvisited scenes~\cite{zhao2025imaginenav, wang2024lookahead,bar2025navigation}. While effective to varying degrees, these approaches share a fundamental limitation in that the agent can only accumulate knowledge from locations it has personally visited or seen. Meanwhile, in real-world settings, multiple robots increasingly operate simultaneously within the same space, such as robot vacuums, surveillance cameras, and personal assistant robots. Yet the VLN community lacks any systematic investigation into whether such incidental inter-agent observation sharing can benefit navigation performance. To bridge this gap, we explore a fundamentally different idea: exploiting the observations of other agents that happen to be navigating in the same environment. It means each agent can supplement its own knowledge with the other's observations, expanding its receptive field without additional exploration (\textit{i.e.}, vision-sharing), as illustrated in Fig.~\ref{fig:intro}. 

We introduce a minimalist evaluation framework named Co-VLN for systematically investigating whether and how other agents' observation benefits VLN. Our framework is driven by a straightforward mechanism: the recognition of spatial overlap. Specifically, when multiple independently navigating agents identify that they have previously traversed the same physical location, regardless of the timing of their visits, they execute a direct exchange of knowledge. By identifying these common spatial anchors, an agent can dynamically integrate the structured memory of its peers into its own state representation, effectively expanding its perceptual field. Importantly, in scenarios where no spatial overlap occurs, our framework naturally reverts to the standard navigation method, leaving the original approach unchanged. This allows our approach to serve as a pure testbed for quantifying the performance gains from shared perception without complex architectural modifications.

To comprehensively evaluate our framework, we conduct extensive experiments on the standard R2R dataset~\cite{anderson2018vision}, instantiating our framework on two representative baselines with distinct paradigms: the learning-based graph navigation model (DUET~\cite{chen2022think}) and the training-free, zero-shot agent (MapGPT~\cite{chen2024mapgpt})\rev{, and additionally to the continuous-environment method ETPNav~\cite{an2024etpnav}}. Remarkably, vision-sharing from peer consistently yields substantial performance improvements across both fundamentally different architectures and paradigms. Beyond the primary results, we conduct in-depth analytical experiments to systematically dissect the dynamics of inter-agent observation sharing, investigating how factors such as MLLM backbones, scene complexity, agent scaling, trajectory pairing strategies, and information sharing strategies influence the gain. Our analysis reveals that the benefits scale favorably with both environment complexity and the number of participating agents, suggesting significant untapped potential in leveraging shared perception across agents for VLN.

In summary, our main contributions are threefold: \textbf{(1)} To the best of our knowledge, we are the first to systematically investigate whether and how inter-agent observation sharing can benefit VLN, demonstrating that independent agents can leverage peer perception to expand their receptive fields without task interference. \textbf{(2)} We propose Co-VLN, a minimalist experimental framework driven by spatial overlap detection, and validate its broad applicability across diverse paradigms, achieving consistent improvements. \textbf{(3)} Through extensive experiments and analyses, we systematically quantify the gains of vision-sharing, establishing a strong foundation and offering valuable insights for future research in collaborative embodied navigation.

\section{Related Work}

\subsection{Vision-Language Navigation}

Vision-Language Navigation~\cite{anderson2018vision} requires an embodied agent to navigate an environment by following natural language instructions. Early methods adopt recurrent architectures for sequential action prediction~\cite{anderson2018vision,fried2018speaker,wang2019reinforced}. Subsequently, pretraining-based methods~\cite{chen2022think,hong2021vln,chen2021history,wang2023scaling,qiao2023hop+} significantly advanced the field, such as DUET~\cite{chen2022think}, which performs dual-scale reasoning over a topological map for both global action planning and local action prediction. The rise of large language models (LLMs) further enabled zero-shot VLN methods~\cite{chen2024mapgpt,zhou2024navgpt,shi2025smartway,zhang2026spatialnav,li2026one} requiring no navigation-specific training, such as MapGPT~\cite{chen2024mapgpt}, which builds an online topological map capturing spatial relationships among explored nodes and prompts GPT to plan multi-step paths over it. More recently, several works leverage large-scale data to train or fine-tune foundation models for VLN~\cite{cheng2024navila,zhang2025embodied,zhang2026qwen}, achieving strong performance. Despite steady progress across these paradigms, all existing methods assume a single navigator operating in isolation, leaving collaboration among concurrent agents in VLN unexplored.

\subsection{Multi-agent Embodied Navigation}

Multi-agent collaboration has been explored to break the limits of individual perception~\cite{wu2025generative}. \rev{Multi-robot navigation has a long history in robotics, spanning coordinated exploration, decentralized collision avoidance, and collaborative mapping~\cite{yan2013survey, burgard2005coordinated,wu2021spatial}. Collaborative SLAM systems such as CCM-SLAM~\cite{schmuck2019ccm}, Kimera-Multi~\cite{tian2022kimera}, and Swarm-SLAM~\cite{lajoie2023swarm} merge metric or metric-semantic maps across robots via inter-robot loop closures. These classical settings pursue geometric or metric objectives with explicit coordination and no language grounding, whereas we focus on the recently emerging setting of semantically-grounded, instruction-following multi-agent navigation.} In VLN, works such as MMCNav~\cite{zhang2025mmcnav} and DiscussNav~\cite{long2024discuss} adopt multi-agent frameworks, but their ``agents'' are multiple LLM roles solving a single navigation task rather than multiple robots. In the broader embodied navigation community, multi-robot collaboration has been studied for image-goal navigation~\cite{wang2021collaborative} and for locating shared~\cite{yu2025co, shen2025enhancing} or multiple~\cite{liu2022multi,rajvanshi2025sayconav} target objects. These methods, however, define success collectively and optimize a single shared objective. In contrast, this paper focuses on VLN, where each agent independently follows its own instruction; rather than optimizing a shared objective, we study whether sharing observations across independently tasked agents mutually improves their individual performance.

\section{VLN with Peer Observation}

\subsection{Problem Formulation}

\subsubsection{Standard VLN}

We briefly revisit the standard Vision-Language Navigation task, typically formulated as a Partially Observable Markov Decision Process (POMDP)~\cite{hao2020towards,liu2024vision}. At each time step $t$, an agent perceives a visual observation $v_t$ from its current viewpoint and follows a natural language instruction $I$. The agent's state $s_t$ is represented by the tuple $(v_t, I, h_t)$, where $h_t$ denotes the navigation history up to time $t$. The agent selects an action $a_t$ from the available action space $\mathcal{A}$ according to its policy $\pi(a_t \mid s_t)$, and transitions to the next state $s_{t+1}$ determined by the transition function $\mathcal{T}(s_{t+1} \mid s_t, a_t)$. The goal is to navigate to a target location specified by $I$.

\begin{figure*}[t]
  \centering
  \includegraphics[width=\textwidth]{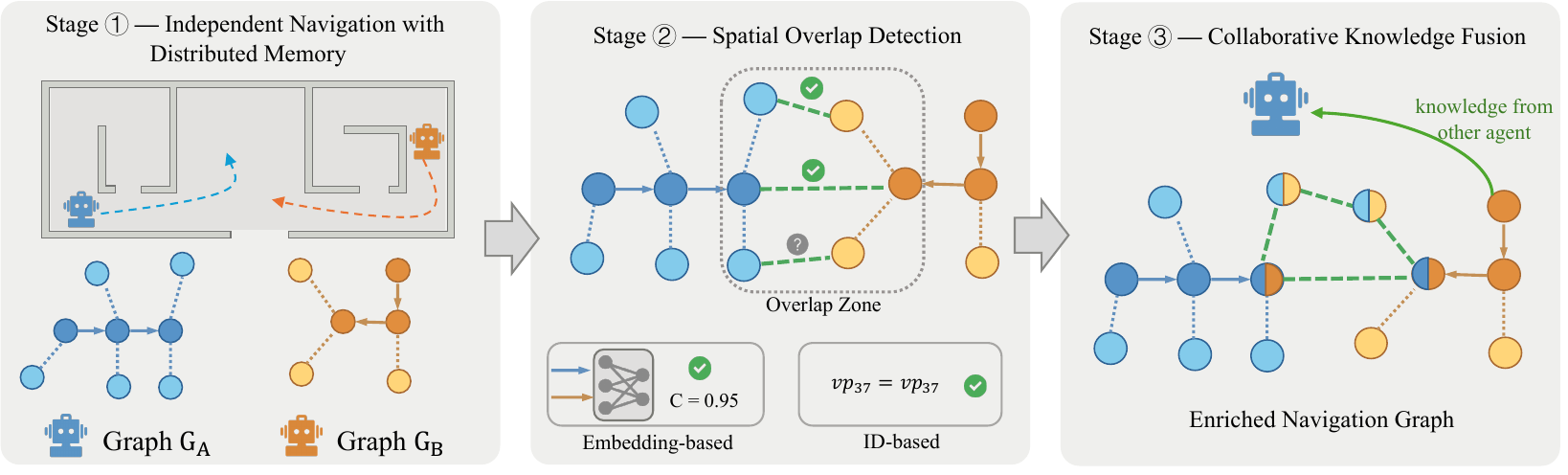}
  \caption{\textbf{Overview of the Co-VLN framework.} Given multiple agents navigating in a shared environment, Co-VLN operates in three stages: \textbf{(1)} each agent independently explores and builds its own topological graph; \textbf{(2)} spatial overlap between the two graphs is detected via embedding-based matching or viewpoint ID matching; \textbf{(3)} the graphs are merged through anchor nodes identified in Stage 2, producing an enriched navigation graph that expands each agent's knowledge beyond its own exploration.}
  \label{fig:framework}
\end{figure*}

\subsubsection{Navigation with Peer Observation}
We extend the standard formulation to a setting where $N$ agents navigate simultaneously within the same environment, each assigned a distinct instruction $I^i$ and heading toward its own independent goal. Each agent $i \in \{1, \dots, N\}$ maintains its own navigation objective and operates its own policy. During navigation, it independently constructs a structured navigational memory $\mathcal{M}_t^i$ (\eg, a topological graph encoding its exploration history). The key distinction from $N$ isolated episodes is the possibility of peer observation sharing. When agents happen to explore overlapping regions, they can exchange their navigational memories. We denote the aggregated memory from all peer agents as $\mathcal{M}_t^{-i} = \{\mathcal{M}_t^j\}_{j \neq i}$. This gives rise to two different policies:
\begin{itemize}
    \item \textbf{Isolated policy} $\pi(a_t^i \mid s_t^i)$, where each agent navigates using only its own observations, equivalent to the original baseline.
    \item \textbf{Vision-sharing policy} $\pi(a_t^i \mid s_t^i, \mathcal{M}_t^{-i})$, where each agent additionally leverages navigational knowledge from its peers.
\end{itemize}
We aim to empirically quantify the performance gap between these two policies across different VLN methods and analyze under what conditions vision-sharing among concurrent agents is beneficial.

\subsection{Co-VLN: A Unified Experimental Framework}

To systematically investigate whether and how inter-agent observation sharing benefits VLN, we design Co-VLN, a unified experimental framework that can be instantiated on top of different VLN methods without modifying their core navigation mechanisms. As illustrated in Fig.~\ref{fig:framework}, Co-VLN decomposes the collaboration process into three sequential stages: (1) independent navigation with distributed memory, where each agent explores the environment autonomously; (2) spatial overlap detection, which identifies whether the agents' explored regions intersect; and (3) collaborative knowledge fusion, which merges navigational knowledge across agents when spatial overlap is detected. We describe each stage below.

\subsubsection{Independent Navigation with Distributed Memory}
Each agent $i$ independently executes its assigned navigation task using a baseline VLN method, following its own instruction $\mathcal{I}^i$. During navigation, each agent incrementally constructs a private navigational memory $\mathcal{M}_t^i$ that encodes its exploration history. This memory takes the form of a topological graph $\mathcal{G}_t^i = (\mathcal{V}_t^i, \mathcal{E}_t^i)$, where nodes $\mathcal{V}_t^i$ represent visited or observed viewpoints and edges $\mathcal{E}_t^i$ capture traversable connections between them. The content stored at each node depends on the navigation model. For example, DUET~\cite{chen2022think} stores learned visual-language embeddings, while some zero-shot methods such as MapGPT~\cite{chen2024mapgpt} store original images or textual scene descriptions. This stage requires no changes to the baseline method, and each agent navigates exactly as it would in the original setting. 

\subsubsection{Spatial Overlap Detection}

The core prerequisite for collaboration is determining whether two agents have observed overlapping regions of the environment. At each time step $t$, we check whether any node in agent $i$'s graph $\mathcal{G}_t^i$ corresponds to a node in agent $j$'s graph $\mathcal{G}_t^j$. If such correspondences exist, we say the agents have achieved spatial intersection, which triggers the subsequent fusion stage. Since node representations differ across methods, the overlap detection mechanism must be instantiated accordingly (Stage~2 of Fig.~\ref{fig:framework}). We describe two instantiations used in our study.

In DUET, each viewpoint is represented as a learned embedding feature. We therefore adopt an embedding-based overlap detection approach, training a lightweight Transformer-based discriminator $f_\theta$ to determine whether two node embeddings correspond to \rev{overlapping observations}. \rev{A naive choice uses an exact-viewpoint label, treating a pair as positive only when both nodes index the same simulator viewpoint. We argue that this target is misaligned with navigation, since a peer is useful when it has explored near a region of interest, not only when it stood on the identical viewpoint. We instead supervise $f_\theta$ with a spatial-overlap target. Let $p_a, p_b \in \mathbb{R}^3$ denote the 3D positions of the two viewpoints, obtained offline from the connectivity graph, and define the training label as:}
\begin{equation}
y(a,b) = \begin{cases} 1 & \text{if } \lVert p_a - p_b \rVert_2 < R,\\ 0 & \text{otherwise,} \end{cases}
\label{eq:overlap}
\end{equation}

\rev{where $R$ denotes the overlap radius, the largest physical distance within which two viewpoints are treated as observing overlapping regions. Positive and negative pairs are mined on the train scenes and the detector is evaluated on the val unseen scenes, which avoids leakage.}

Given a pair of node embeddings $(e_a, e_b)$ where $e_a \in \mathcal{V}_t^i$ and $e_b \in \mathcal{V}_t^j$, the discriminator outputs a confidence score $c$:
\begin{equation}
c = f_\theta(e_a, e_b) \in [0, 1].
\end{equation}
\rev{This score indicates the likelihood that the two nodes are spatially co-located within radius $R$.} When $c$ exceeds a threshold $\tau$, \rev{set to $0.5$ in our experiments, we treat the two nodes as a candidate matched pair}. To enable smooth integration in the subsequent fusion stage, we further convert the confidence score into an estimated inter-node distance: 
\begin{equation}
d(e_a, e_b) = (1 - c) \cdot \alpha, 
\end{equation}
where $\alpha=5$ is a scaling factor. This formulation ensures that high-confidence matches yield near-zero distances, allowing the fused graph to maintain geometrically coherent connectivity. 

In MapGPT, the topological graph is originally constructed using simulator-provided viewpoint IDs, so we directly match nodes across agents by comparing these identifiers. We refer to this as the ID-based overlap detection approach. No additional information is introduced beyond what the baseline method already uses.

\subsubsection{Collaborative Knowledge Fusion}

Once spatial overlap is detected, the agents merge their navigational graphs to form an enriched representation. As depicted in Stage~3 of Fig.~\ref{fig:framework}, for each agent $i$, we augment its private graph $\mathcal{G}_t^i$ at current time step with nodes and edges from agent $j$'s graph $\mathcal{G}_t^j$ that are not already present in $\mathcal{G}_t^i$. Matched node pairs identified in the previous stage serve as bridges that connect the two subgraphs. Formally, the fused graph for agent $i$ is constructed as:
\begin{equation}
  \tilde{\mathcal{G}}_t^i = \big(\mathcal{V}_t^i \cup \mathcal{V}_t^{j \setminus i},\; \mathcal{E}_t^i \cup \mathcal{E}_t^{j \setminus i} \cup \mathcal{E}_t^{\mathrm{bridge}}\big),
  \label{eq:fusion}
\end{equation}
where $\mathcal{V}_t^{j \setminus i}$ denotes the set of nodes in agent $j$'s graph that have no match in agent $i$'s graph, $\mathcal{E}_t^{j \setminus i}$ denotes the corresponding edges, and $\mathcal{E}_t^{\mathrm{bridge}}$ are newly created edges connecting the selected matched node pairs across the two graphs\rev{, whose selection is detailed below}. For embedding-based detection, the bridge edge weights are derived from the estimated distances $d(e_a, e_b)$ in the overlap detection stage. For ID-based detection, matched nodes refer to the identical physical location, so they are directly merged with zero distance.

\rev{For embedding-based detection, multiple candidate matched pairs sometimes emerge. Let $\{(a_k, b_k)\}$, with $a_k \in \mathcal{G}_t^i$ and $b_k \in \mathcal{G}_t^j$, denote the candidate matched pairs that exceed the confidence threshold, ordered by confidence so that $(a_1, b_1)$ is the highest-confidence pair, which we take as the anchor. Using only the anchor is robust but conservative, whereas admitting additional bridges localizes a larger peer subgraph but risks false bridges from detector false positives that distort the fused graph~\cite{eysenbach2019search}. To admit additional bridges safely, we apply a geometry-consistency verification that requires no shared absolute coordinates. If a candidate bridge $(a_k, b_k)$ and the anchor $(a_1, b_1)$ are both correct, the geodesic distance between their endpoints must agree across the two graphs:}
\begin{equation}
\big| d_{\mathcal{G}^i}(a_1, a_k) - d_{\mathcal{G}^j}(b_1, b_k) \big| < \tau_{g},
\label{eq:geom}
\end{equation}
\rev{where $d_{\mathcal{G}^i}$ and $d_{\mathcal{G}^j}$ are distances computed within agent $i$'s and agent $j$'s own graphs, respectively, and $\tau_g$ is a distance tolerance, set to $1\text{m}$. Starting from the anchor, we admit each further candidate that satisfies this condition. This filter recovers naive multi-bridging from collapse and prunes false bridges.}

After fusion, each agent continues with its original policy, but now operates over the enriched graph $\tilde{\mathcal{G}}_t^i$ rather than its private graph $\mathcal{G}_t^i$. The agent can thus consider frontiers and paths that its partner discovered but it has not personally explored. The fusion is symmetric, so both agents receive the other's knowledge. Because they follow different instructions toward different goals, however, the same fused information may be leveraged differently, and an agent benefits most when its partner's observations cover regions relevant to its own goal.

\subsection{Task Construction Protocol}

\label{sec:task_construction}

Existing VLN benchmarks evaluate each episode in isolation, so we construct concurrent episodes in which multiple agents navigate the same environment simultaneously. Given a dataset with $K$ episodes, we partition them into $K/N$ groups of $N$ episodes sharing the same scene, and within each group the $N$ agents navigate in parallel, each following its own instruction. Metrics are computed per-episode and averaged over all $K$ episodes, so the evaluation covers exactly the same task set as the single-agent baseline and introduces no advantage beyond observation sharing. Each agent keeps the same instruction, start position, and model, differing only in that it may receive peer graph information. If no overlap occurs during an episode, its behavior is identical to the baseline.

How episodes are grouped affects the chance of peer assistance, so we consider two strategies. \textbf{Prior-based pairing} pairs episodes whose ground-truth trajectories share overlapping viewpoints, maximizing the chance that agents traverse common regions. This mirrors a household setting where robots pass through shared spaces such as hallways and living rooms even under different tasks. \textbf{Random pairing} instead groups episodes within a scene without regard to spatial correlation, testing whether peer observation still helps under less favorable conditions.

\section{Experiments}
\subsection{Experimental Setup}

\subsubsection{Dataset and Evaluation Metrics}
We conduct experiments on the Room-to-Room (R2R) dataset~\cite{anderson2018vision}, the most widely used benchmark for VLN. R2R is built upon the Matterport3D~\cite{Matterport3D}, which provides photo-realistic indoor environments. This dataset comprises 21,567 instructions across 90 building-scale scenes, and we primarily evaluated on the val unseen split, featuring 2,349 instruction-trajectory pairs across 11 scenes. We evaluate navigation performance by using standard VLN metrics, including Trajectory Length (TL), Navigation Error (NE), Oracle Success Rate (OSR), Success Rate (SR), and Success weighted by Path Length (SPL).

\subsubsection{Baseline Methods}

To demonstrate the generality of the inter-agent observation sharing setting, we adopt two representative VLN methods from different paradigms as our baselines: \textbf{(1) DUET}~\cite{chen2022think}, a supervised learning method that maintains a topological map and performs dual-scale action prediction, representative of the pre-training and fine-tuning paradigm; and \textbf{(2) MapGPT}~\cite{chen2024mapgpt}, a zero-shot approach that leverages multimodal large language models (MLLMs) for navigation decisions, constructing a textual topological map as spatial memory. It represents the emerging paradigm of training-free agents. By building upon these two fundamentally different architectures, we aim to demonstrate that the benefits of peer observation sharing are paradigm-agnostic.

\subsubsection{Episode Pairing Setup}
Following the task construction protocol in Sec.~\ref{sec:task_construction}, we evaluate under the two-agent setting ($N=2$) as the default configuration. For R2R val unseen, the $K=2349$ episodes are partitioned into pairs within each scan, where paired episodes must correspond to different trajectories. A small number of episodes (5 out of 2349) cannot be paired because certain scans contain an odd number of trajectories, and are instead duplicated as self-pairs, effectively reducing to the original baseline. Under prior-based pairing, we select pairs whose ground-truth trajectories are spatially proximate, resulting in an average of $4.24$ overlapping viewpoints per pair. Under random pairing, this number drops to $0.88$. Unless otherwise specified, most experiments in this paper adopt the prior-based pairing strategy.

\subsubsection{Implementation Details}
For DUET, we initialize from the officially released weights and fine-tune under our setting with a batch size of 8 for 20,000 iterations. For MapGPT, we adopt Qwen3-VL-32B~\cite{bai2025qwen3} as the default MLLM backbone, and retain the original prompting protocol with only minimal modifications to incorporate the necessary vision-sharing context. All other hyperparameters and training configurations follow the respective original baselines without modification. Experiments are conducted using $2 \times$ NVIDIA A6000 GPUs.

\begin{table*}[t]
\centering
\caption{Comparison with state-of-the-art methods on the R2R val unseen.
$\uparrow$: higher is better, $\downarrow$: lower is better. MapGPT$^{\ddagger}$ uses Qwen3-VL-32B as the MLLM. $\dagger$ denotes the use of oracle viewpoint IDs for spatial overlap detection. $*$ denotes applying CLIP ViT-B16 as image features. Best results are \textbf{bold}.}
\label{tab:main_results}
\setlength{\tabcolsep}{4pt}
\resizebox{\linewidth}{!}{%
\begin{tabular}{llccccc}
\toprule
\textbf{\shortstack{Method\\Paradigm}} & \textbf{Methods} & \textbf{TL} & \textbf{NE$\downarrow$} & \textbf{OSR$\uparrow$} & \textbf{SR$\uparrow$} & \textbf{SPL$\uparrow$} \\
\midrule
& PREVALENT~\cite{hao2020towards}          & 10.19 & 4.71 & --   & 58   & 53 \\
& VLN\,$\circlearrowright$\,BERT~\cite{hong2021vln} & 12.01 & 3.93 & 69 & 63 & 57 \\
& HAMT~\cite{chen2021history}               & 11.46 & 3.62 & 73   & 66   & 61 \\
Supervised & HOP+~\cite{qiao2023hop+}               & 11.76 & 3.49 & --   & 67   & 61 \\
Learning & TD-STP~\cite{zhao2022target}             & -- & 3.22 & 76   & 70   & 63 \\
& BEVBert~\cite{an2023bevbert}            & 14.55 & 2.81 & 84 & 75 & 64 \\
\cmidrule{2-7}
\rowcolor{blue!8}
& DUET~\cite{chen2022think} (baseline)    & 13.94 & 3.31 & 80.54 & 71.52 & 60.41 \\
\rowcolor{blue!8}
& DUET + Vision-Sharing        &  13.60 &  2.99  &  82.72   & 74.50 & 62.56 \\
\rowcolor{blue!8}
& DUET + Vision-Sharing$^{\dagger}$        & 14.62 & 2.63   & 85.06   & 76.23 & 62.90 \\
\rowcolor{blue!8}
& DUET$^{*}$+ScaleVLN~\cite{wang2023scaling}    & 12.40 & 2.38 & 87.48 & 79.40 & 69.97 \\
\rowcolor{blue!8}
& DUET$^{*}$+ScaleVLN + Vision-Sharing        & 12.44 & 2.15  & 88.85   & 81.01 & 72.39 \\
\rowcolor{blue!8}
& DUET$^{*}$+ScaleVLN + Vision-Sharing$^{\dagger}$        & 12.31 & \textbf{2.12}   & \textbf{89.10}   & \textbf{82.46} & \textbf{73.39} \\
\midrule
& NavGPT~\cite{zhou2024navgpt}             & 11.45 & 6.46 & 42   & 34   & 29 \\
& DiscussNav~\cite{long2024discuss}    & 9.69 & 5.32 & 61.0 & 43.0 & 40.0 \\
& MapGPT~\cite{chen2024mapgpt}    & -- & 5.63 & 57.6 & 43.7 & 34.8 \\
Zero-Shot & MC-GPT~\cite{zhan2024mc}             & -- & 5.42 & 68.8 & 32.1 & -- \\
& MSNav~\cite{liu2025msnav}              & -- & 5.24 & 65.0   & 46.0   & 40.0 \\
& SpatialGPT~\cite{jiang2025spatialgpt}         & -- & 5.56 & \textbf{70.8} & 48.4 & 36.1 \\
\cmidrule{2-7}
\rowcolor{blue!8}
& MapGPT$^{\ddagger}$ (baseline) & 12.16 & 4.80 & 63.98 & 52.19 & 44.73 \\
\rowcolor{blue!8}
& MapGPT$^{\ddagger}$ + Vision-Sharing$^{\dagger}$ & 12.20 & \textbf{4.36} & 68.58 & \textbf{55.81} & \textbf{47.26} \\
\bottomrule
\end{tabular}}
\end{table*}

\subsection{Main Results}

To investigate whether peer observation sharing fundamentally benefits VLN, we compare our vision-sharing setting against state-of-the-art VLN methods on the R2R val unseen split. As shown in Table~\ref{tab:main_results}, incorporating vision-sharing into established baselines yields consistent improvements across different navigation paradigms.

Under the supervised learning paradigm, applying vision-sharing to DUET yields substantial improvements across all metrics. With embedding-based overlap detection, DUET + Vision-Sharing raises SR from $71.52\%$ to $74.50\%$ ($+2.98$) and SPL from $60.41\%$ to $62.56\%$ ($+2.15$). When using oracle viewpoint IDs for overlap detection (an upper-bound setting), SR further improves to $76.23\%$ and SPL to $62.90\%$, suggesting that a more accurate spatial overlap detector could unlock additional performance gains. We further apply vision-sharing to DUET+ScaleVLN~\cite{wang2023scaling}, the current top-performing supervised method that builds upon DUET with large-scale augmented training data, and observe consistent improvements, establishing a new state-of-the-art on R2R val unseen.

In the zero-shot paradigm, vision-sharing brings improvements to MapGPT as well, improving SR from $52.19\%$ to $55.81\%$ ($+3.62$) and SPL from $44.73\%$ to $47.26\%$ ($+2.53$), while NE decreases from $4.80$ to $4.36$. The resulting model achieves the best SR and SPL among the compared zero-shot methods. Overall, the consistent gains across both a learned graph model and a zero-shot MLLM agent confirm that peer observation sharing generalizes across paradigms.

\subsection{Experiments on Continuous Environments}
\begin{table}[t]
\centering
\caption{Comparison with state-of-the-art methods on the R2R-CE val unseen.
$\uparrow$: higher is better, $\downarrow$: lower is better. Best results are \textbf{bold}.}
\label{tab:ce_results}
\setlength{\tabcolsep}{6pt}
\renewcommand{\arraystretch}{1.1}
\resizebox{0.8\linewidth}{!}{%
\begin{tabular}{lccccc}
\toprule
\textbf{Methods} & \textbf{TL} & \textbf{NE$\downarrow$} & \textbf{OSR$\uparrow$} & \textbf{SR$\uparrow$} & \textbf{SPL$\uparrow$} \\
\midrule
Seq2Seq~\cite{krantz2020beyond}        & 8.64  & 7.37 & 40 & 32 & 30 \\
HPN~\cite{krantz2021waypoint}            & 7.62  & 6.31 & 40 & 36 & 34 \\
CM2~\cite{georgakis2022cross}            & 11.54 & 7.02 & 42 & 34 & 28 \\
WS-MGMAP~\cite{chen2022weakly}       & 10.00 & 6.28 & 48 & 39 & 34 \\
CWP-CMA~\cite{hong2022bridging}       & 10.90 & 6.20 & 52 & 41 & 36 \\
CWP-RecBERT~\cite{hong2022bridging}    & 12.23 & 5.74 & 53 & 44 & 39 \\
Sim2Sim~\cite{krantz2022sim}        & 10.69 & 6.07 & 52 & 43 & 36 \\
BEVBert~\cite{an2023bevbert}        & -- & 4.57 & \textbf{67} & 59 & 50 \\
\cmidrule{1-6}
\rowcolor{blue!8}
ETPNav~\cite{an2024etpnav} (baseline)       & 11.99 & 4.71 & 64.71 & 57.21 & 49.15 \\
\rowcolor{blue!8}
ETPNav + Vision-Sharing & 10.56 & \textbf{4.66} & 64.71 & \textbf{59.16} & \textbf{51.79} \\
\bottomrule
\end{tabular}}
\end{table}

\rev{To validate generalizability beyond discrete navigation graphs, we extend our framework to the continuous setting using R2R-CE~\cite{krantz2020beyond}, which ports the R2R~\cite{anderson2018vision} instructions into the Habitat~\cite{habitat19iccv} simulator, where agents act with low-level controls in continuous 3D space rather than jumping between predefined viewpoints. We adopt ETPNav~\cite{an2024etpnav}, one of the state-of-the-art continuous-VLN methods.}

\rev{Since agents operate in a continuous coordinate space, we use GPS coordinates for overlap detection, triggering graph fusion when two agents' positions fall within $0.5\,\text{m}$. On the R2R-CE val unseen split ($1{,}839$ episodes) under prior-based pairing, fine-tuning ETPNav within our framework improves SR from $57.21\%$ to $59.16\%$ ($+1.95$) and SPL from $49.15\%$ to $51.79\%$ ($+2.64$) (Table~\ref{tab:ce_results}), confirming that peer observation sharing generalizes to continuous environments.}

\subsection{Scalability and Generalization}

Having established the effectiveness of peer observation sharing, we further investigate its scalability along two dimensions: the number of concurrent agents and the capability of different foundation models. To amplify observable differences and reduce evaluation cost, we conduct the following experiments on a challenging subset we term R2R-Hard Subset, consisting of 200 episodes sampled from one of the most difficult scans in the R2R val unseen split, where both the baseline and our method attain low SR and SPL.

\begin{figure*}[t]
\centering
\begin{minipage}[t]{0.4\textwidth}
    \centering
    \vspace{0pt}
    \includegraphics[width=\linewidth]{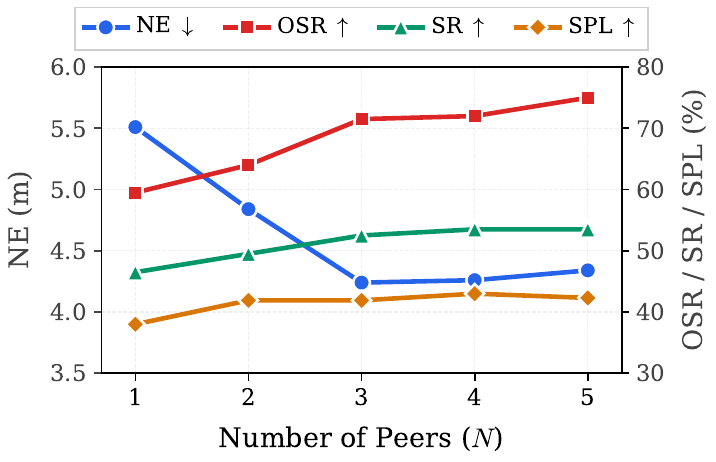}
    \captionof{figure}{Peer scaling trends.}
    \label{fig:scaling}
\end{minipage}\hfill
\begin{minipage}[t]{0.57\textwidth}
    \centering
    \vspace{0pt}
    \captionof{table}{Effect of scaling the number of concurrent peers. Best results are \textbf{bold}, second best are \underline{underlined}.}
    \label{tab:num_agents}
    \setlength{\tabcolsep}{4pt}
    \resizebox{\linewidth}{!}{%
    \begin{tabular}{lccccc}
    \toprule
    \textbf{Peers ($N$)} & \textbf{Vision-Sharing} & \textbf{NE$\downarrow$} & \textbf{OSR$\uparrow$} & \textbf{SR$\uparrow$} & \textbf{SPL$\uparrow$} \\
    \midrule
    1 (baseline) & \ding{55} & 5.51 & 59.5 & 46.5 & 38.0 \\
    2 & \ding{51} & 4.84 & 64.0 & 49.5 & 41.9 \\
    3 & \ding{51} & \textbf{4.24} & 71.5 & \underline{52.5} & 41.9 \\
    4 & \ding{51} & \underline{4.26} & \underline{72.0} & \textbf{53.5} & \textbf{43.0} \\
    5 & \ding{51} & 4.34 & \textbf{75.0} & \textbf{53.5} & \underline{42.3} \\
    \bottomrule
    \end{tabular}}
\end{minipage}
\end{figure*}

\definecolor{gain}{rgb}{0.0, 0.5, 0.0}
\begin{table}[t]
\centering
\caption{Effect of different MLLMs on Vision-Sharing setting. Best results are \textbf{bold}, second best are \underline{underlined}.
\textsuperscript{$\dagger$} indicates closed-source models.}
\label{tab:mllm_ablation}
\setlength{\tabcolsep}{3pt}
\resizebox{\linewidth}{!}{%
\begin{tabular}{l c cccc}
\toprule
\textbf{MLLM} & \textbf{\shortstack{Vision-\\Sharing}} & \textbf{NE$\downarrow$} & \textbf{OSR$\uparrow$} & \textbf{SR$\uparrow$} & \textbf{SPL$\uparrow$} \\
\midrule
\multirow{2}{*}{InternVL3.5-38B}
 & \ding{55} & 6.60 & 50.5 & 37.0 & 31.1 \\
 & \ding{51} & 6.40 {\footnotesize\textcolor{gain}{(-0.20)}} & 51.5 {\footnotesize\textcolor{gain}{(+1.0)}} & 38.0 {\footnotesize\textcolor{gain}{(+1.0)}} & 31.6 {\footnotesize\textcolor{gain}{(+0.5)}} \\
\midrule
\multirow{2}{*}{Gemini-3-Flash\textsuperscript{$\dagger$}}
 & \ding{55} & 6.09 & 40.5 & 38.0 & 35.2 \\
 & \ding{51} & 5.10 {\footnotesize\textcolor{gain}{(-0.99)}} & 59.0 {\footnotesize\textcolor{gain}{(+18.5)}} & 46.0 {\footnotesize\textcolor{gain}{(+8.0)}} & 40.5 {\footnotesize\textcolor{gain}{(+5.3)}} \\
\midrule
\multirow{2}{*}{Qwen3-VL-32B}
 & \ding{55} & 5.51 & 59.5 & 46.5 & 38.0 \\
 & \ding{51} & 4.84 {\footnotesize\textcolor{gain}{(-0.67)}} & 64.0 {\footnotesize\textcolor{gain}{(+4.5)}} & 49.5 {\footnotesize\textcolor{gain}{(+3.0)}} & 41.9 {\footnotesize\textcolor{gain}{(+3.9)}} \\
\midrule
\multirow{2}{*}{GPT-5.2\textsuperscript{$\dagger$}}
 & \ding{55} & 4.35 & 67.5 & 54.5 & 46.8 \\
 & \ding{51} & \textbf{3.47} {\footnotesize\textcolor{gain}{(-0.88)}} & \underline{71.5} {\footnotesize\textcolor{gain}{(+4.0)}} & \underline{60.0} {\footnotesize\textcolor{gain}{(+5.5)}} & \underline{50.1} {\footnotesize\textcolor{gain}{(+3.3)}} \\
\midrule
\multirow{2}{*}{Gemini-2.5-Pro\textsuperscript{$\dagger$}}
 & \ding{55} & 4.28 & 68.5 & 56.0 & 49.1 \\
 & \ding{51} & \underline{3.76} {\footnotesize\textcolor{gain}{(-0.52)}} & \textbf{74.5} {\footnotesize\textcolor{gain}{(+6.0)}} & \textbf{64.5} {\footnotesize\textcolor{gain}{(+8.5)}} & \textbf{53.8} {\footnotesize\textcolor{gain}{(+4.7)}} \\
\bottomrule
\end{tabular}%
}
\end{table}

\subsubsection{Scaling the Number of Concurrent Peers}

To study how the benefit scales with the number of peers, we evaluate MapGPT on the R2R-Hard Subset by scaling the number of agents ($N$) from $1$ to $5$. Note that unlike the episode-pairing protocol used in the main experiments, here we retain all 200 episodes as primary tasks and assign each one $N-1$ peer agents with maximally overlapping ground-truth paths. This design aims to reveal the best-case performance gain as the number of peers increases.

As shown in Table~\ref{tab:num_agents} and Fig.~\ref{fig:scaling}, adding more peers consistently improves navigation performance up to a saturation point. SR increases steadily from $46.5\%$ ($N\!=\!1$) to $53.5\%$ ($N\!=\!4$), with the largest gains occurring between $N\!=\!1$ and $N\!=\!3$. Beyond $N\!=\!4$, performance saturates. At $N\!=\!5$, SR remains at $53.5\%$ while SPL and NE slightly degrade, suggesting that excessive peers introduce redundant information that marginally harms efficiency. This is expected, as the probability that an additional peer provides novel spatial coverage decreases with more agents already exploring. In practice, $N\!=\!2$ offers the best trade-off between performance gain and resource expenditure, aligning well with realistic household scenarios where $2$--$3$ robots may coexist.

\subsubsection{Generalization Across MLLMs}

To verify generalization across different foundation models, we evaluate MapGPT using five MLLMs of varying capacities. These include open-source models (InternVL3.5-38B~\cite{wang2025internvl3} and Qwen3-VL-32B~\cite{bai2025qwen3}) as well as state-of-the-art closed-source models (Gemini-3-Flash~\cite{deepmind2025gemini3flash}, Gemini-2.5-Pro~\cite{comanici2025gemini}, and GPT-5.2~\cite{openai2025gpt52}). 

We conduct this evaluation on R2R-Hard Subset. As shown in Table~\ref{tab:mllm_ablation}, vision-sharing improves performance across all tested MLLMs, although the gain varies widely ($+1.0$ to $+8.5$ SR). This variance is not explained by overall model strength. InternVL3.5-38B and Gemini-3-Flash start from nearly identical baselines ($37.0\%$ vs.\ $38.0\%$ SR), yet gain $+1.0$ and $+8.0$ respectively, while GPT-5.2 gains less than both Gemini models. We attribute this to differences in reasoning over larger graphs, as vision-sharing substantially enlarges the topological map provided to the MLLM. Models better at reasoning over larger graphs more effectively convert the expanded coverage into correct decisions.

\subsection{Analysis Experiments}

We now conduct a series of analyses to understand when and how vision-sharing benefits VLN, focusing on scene area, trajectory pairing strategies, and graph  fusion strategy.

\subsubsection{Effect of Scene Area}
Intuitively, navigating a large, complex house is far more prone to error than crossing a small apartment. To quantify how environment size affects the benefit of peer vision-sharing, we categorize the R2R val unseen scenes into three groups based on their navigable area: Small ($<250\,\text{m}^2$), Medium ($250\text{--}450\,\text{m}^2$), and Large ($>450\,\text{m}^2$).

\newcommand{\del}[2]{#1\,\textcolor{gray}{\footnotesize{(#2)}}}
\newcommand{\bestdel}[2]{\textbf{#1}\,\textbf{\footnotesize{(#2)}}}

\begin{table}[t]
\centering
\caption{Effect of scene area on Vision-Sharing setting. Shaded rows show absolute improvement with relative improvement (\%). \textbf{Bold} indicates the best improvement.}
\label{tab:area_impact}
\setlength{\tabcolsep}{3pt}
\renewcommand{\arraystretch}{1.15}
\resizebox{\linewidth}{!}{%
\begin{tabular}{l c cccc}
\toprule
\textbf{Area Group} & \textbf{\shortstack{Vision-\\Sharing}} & \textbf{SR} $\uparrow$ & \textbf{SPL} $\uparrow$ & \textbf{OSR} $\uparrow$ & \textbf{NE} $\downarrow$ \\
\midrule
\multirow{2}{*}{Small ($<$250\,m$^2$)}
 & \ding{55} & 56.75 & 49.95 & 68.65 & 3.77 \\
 & \ding{51} & 58.73 & 50.71 & 72.62 & 3.62 \\
\rowcolor{gray!10} \multicolumn{2}{r}{} & \del{$+$1.98}{$+$3.49\%} & \del{$+$0.76}{$+$1.52\%} & \del{$+$3.97}{$+$5.78\%} & \del{$-$0.15}{$-$3.98\%} \\
\midrule
\multirow{2}{*}{Medium (250--450\,m$^2$)}
 & \ding{55} & 52.40 & 43.62 & 61.74 & 4.26 \\
 & \ding{51} & 55.81 & 46.75 & 67.30 & 3.89 \\
\rowcolor{gray!10} \multicolumn{2}{r}{} & \del{$+$3.41}{$+$6.51\%} & \bestdel{$+$3.13}{$+$7.18\%} & \bestdel{$+$5.56}{$+$9.01\%} & \del{$-$0.37}{$-$8.69\%} \\
\midrule
\multirow{2}{*}{Large ($>$450\,m$^2$)}
 & \ding{55} & 49.86 & 43.07 & 63.44 & 5.69 \\
 & \ding{51} & 54.42 & 45.99 & 67.62 & 5.06 \\
\rowcolor{gray!10} \multicolumn{2}{r}{} & \bestdel{$+$4.56}{$+$9.15\%} & \del{$+$2.92}{$+$6.78\%} & \del{$+$4.18}{$+$6.59\%} & \bestdel{$-$0.63}{$-$11.07\%} \\
\midrule
\midrule
\multirow{2}{*}{\textbf{Overall}}
 & \ding{55} & 52.19 & 44.73 & 63.98 & 4.80 \\
 & \ding{51} & 55.81 & 47.26 & 68.58 & 4.36 \\
\rowcolor{gray!10} \multicolumn{2}{r}{} & \del{$+$3.62}{$+$6.94\%} & \del{$+$2.53}{$+$5.66\%} & \del{$+$4.60}{$+$7.19\%} & \del{$-$0.44}{$-$9.17\%} \\
\bottomrule
\end{tabular}%
}
\end{table}

As shown in Table~\ref{tab:area_impact}, we evaluate MapGPT under this categorization and find that the benefits of peer observation exhibit a strong positive correlation with scene size. In small environments, the vision-sharing yields a modest $+1.98$ absolute improvement in SR. However, in medium and large environments, the SR gains surge to $+3.41$ and $+4.56$, respectively. Similarly, large scenes show the largest drop in NE ($-0.63\,$ compared to $-0.15\,$ in small scenes). This trend validates our motivation, as agents in larger spaces face more decision points and are more prone to choosing wrong directions, a peer agent effectively serves as a concurrent explorer, expanding the shared perceptual horizon and preventing the primary agent from getting lost during long navigation tasks.

\subsubsection{Effect of Pairing Strategy}
\label{sec:pairing_analysis}

While our default prior-based pairing maximizes overlap, such prior knowledge is unavailable in real deployment, so we compare it against random pairing, where episodes are paired without any spatial considerations.

Table~\ref{tab:split_strategy} presents the results on the R2R val unseen split. On both DUET and MapGPT, random pairing still outperforms the original baseline across all metrics, confirming that peer observation sharing provides benefits even without careful task assignment. This is because, even under random pairing, agents navigating the same environment have a chance of exploring overlapping regions. However, employing the prior-based pairing can unlock the full potential of the framework, further boosting DUET's SR to $74.5\%$ and MapGPT's SR to $55.8\%$.

\begin{table}[t]
\centering
\centering
\caption{Effect of trajectory pairing strategies on Vision-Sharing setting. Best results are \textbf{bold}, second best are \underline{underlined}.}
\label{tab:split_strategy}
\setlength{\tabcolsep}{4pt}
\resizebox{0.8\linewidth}{!}{%
\begin{tabular}{l c c cccc}
\toprule
\textbf{Baseline} & \textbf{\shortstack{Vision-\\Sharing}} & \textbf{\shortstack{Pairing\\Strategy}} & \textbf{NE$\downarrow$} & \textbf{OSR$\uparrow$} & \textbf{SR$\uparrow$} & \textbf{SPL$\uparrow$} \\
\midrule
\multirow{3}{*}{MapGPT}
 & \ding{55} & -- & 4.80 & 64.0 & 52.2 & 44.7 \\
 & \ding{51} & random pairing & \underline{4.66} & \underline{65.3} & \underline{53.4} & \underline{45.7} \\
 & \ding{51} & prior-based pairing & \textbf{4.36} & \textbf{68.6} & \textbf{55.8} & \textbf{47.3} \\
\midrule
\multirow{3}{*}{DUET}
 & \ding{55} & -- & 3.31 & 80.5 & 71.5 & 60.4 \\
 & \ding{51} & random pairing & \underline{3.24} & \underline{81.2} & \underline{72.6} & \underline{61.2} \\
 & \ding{51} & prior-based pairing & \textbf{2.99} & \textbf{82.7} & \textbf{74.5} & \textbf{62.6} \\
\bottomrule
\end{tabular}}
\end{table}

\begin{table}[t]
\centering
\caption{Effect of graph fusion strategy. Best results are \textbf{bold}, second best are \underline{underlined}.}
\label{tab:merge_timing}
\setlength{\tabcolsep}{4pt}
\resizebox{0.65\linewidth}{!}{%
\begin{tabular}{ccc cc}
\toprule
\textbf{\shortstack{Detection\\Trigger}} & \textbf{\shortstack{Bi-\\direction}} & \textbf{Persistent} & \textbf{SR$\uparrow$} & \textbf{SPL$\uparrow$} \\
\midrule
\multicolumn{3}{c}{DUET (w/o Vision-Sharing)} & 71.5 & 60.4 \\
\midrule
\checkmark & \checkmark & $\times$ & 72.4 & 60.0 \\
\checkmark & $\times$ & \checkmark & 73.7 & 61.3 \\
$\times$ & \checkmark & \checkmark & \underline{74.3} & \underline{61.5} \\
\checkmark & \checkmark & \checkmark & \textbf{74.5} & \textbf{62.6} \\
\bottomrule
\end{tabular}}
\end{table}

\subsubsection{Effect of Graph Fusion Strategy}
Finally, we study how different graph fusion strategies affect performance. We conduct this ablation on vanilla DUET to avoid interference from other factors. We examine three dimensions: (1) Trigger: whether the sharing is initiated as soon as the overlap detector identifies that two nodes from different agents correspond to the same location (Detection Trigger), or only after both agents have physically visited the same viewpoint; (2) Direction: whether sharing information flows to both agents (Bi-direction) or only to the later-arriving agent; and (3) Persistent: whether the sharing continues after the initial trigger or occurs only once.

Table~\ref{tab:merge_timing} reveals that enabling all three options yields the best performance, substantially outperforming the original baseline. Altering any single design choice leads to a performance drop. From these results, we draw a clear conclusion that merging early, broadly, and continuously maximizes the benefit of peer observation sharing. Agents should share information as soon as their observations overlap, in both directions, and continue sharing throughout the remainder of navigation.

\subsection{Qualitative Analysis}

\begin{figure*}[t]
  \centering
  \includegraphics[width=\textwidth]{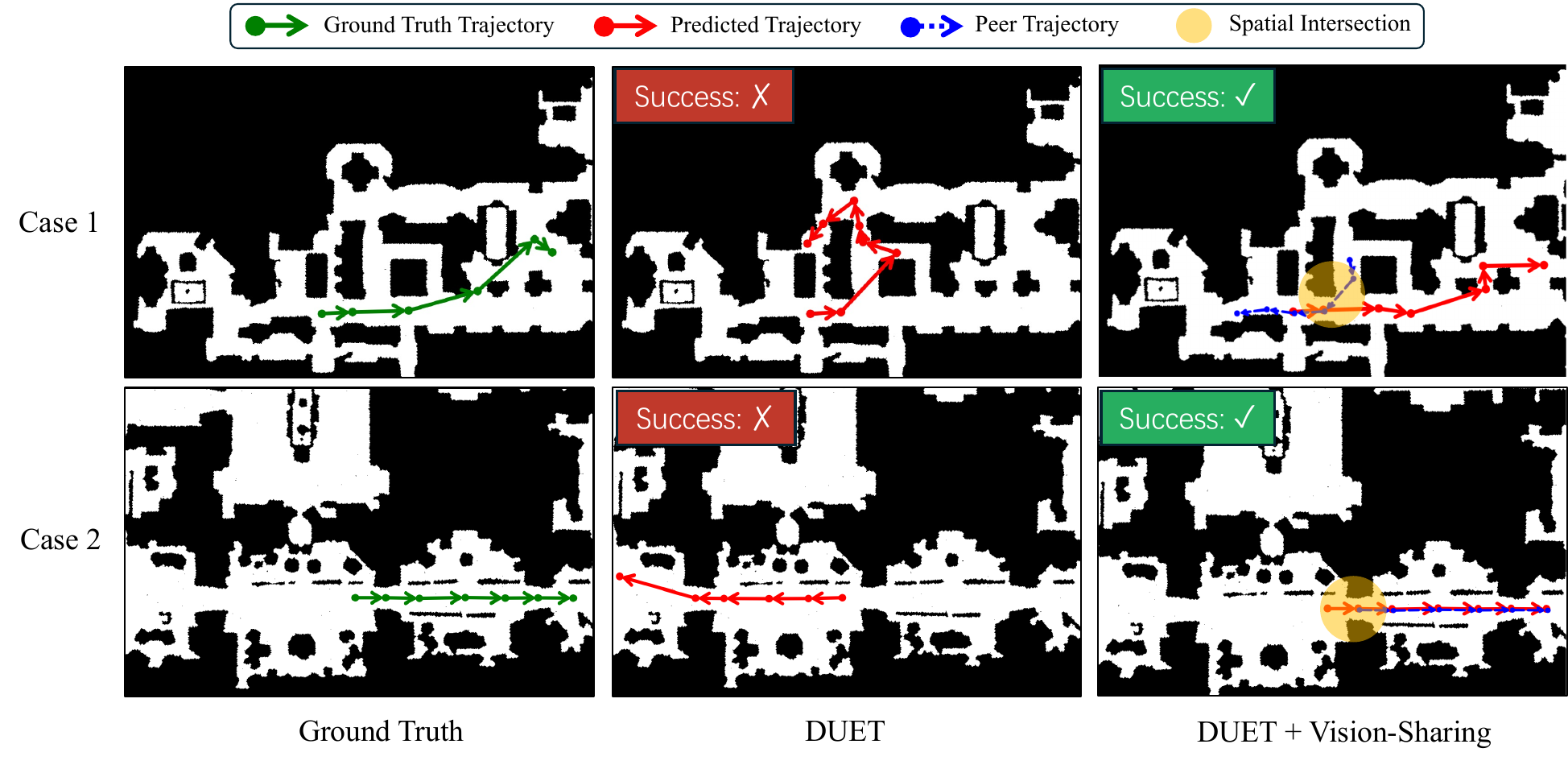}
 \caption{\textbf{Qualitative comparison}. We visualize DUET with and without vision-sharing.}
  \label{fig:vis}
\end{figure*}

Fig.~\ref{fig:vis} presents two representative cases where vision-sharing transforms navigation failure into success. In Case~1, the agent without peer observation deviates upward into irrelevant rooms. With peer observation, it recognizes that the upward region does not match its instruction and navigates toward the correct path instead. In Case~2, the original DUET navigates in the completely opposite direction (leftward). Peer observations reveal the spatial structure to the right, redirecting the agent toward the goal. Both cases illustrate that peer observations help reduce uncertainty at critical decision points, preventing early commitment to wrong directions.

\section{Conclusion}

In this work, we introduce a new perspective for VLN research by investigating whether peer observations from concurrently navigating agents can benefit navigation performance. To study this question, we propose Co-VLN, a minimalist experimental framework, and instantiate it on two representative VLN methods spanning different paradigms: the supervised learning-based DUET and the zero-shot MapGPT. Experimental results demonstrate that vision sharing from peer agents substantially improves VLN performance across both paradigms\rev{, and further generalizes to continuous environments with ETPNav}. Through extensive analyses, we further reveal when and how such vision-sharing is most effective. We hope our findings provide a solid foundation for future research on collaborative embodied navigation.

\clearpage
\appendix
\renewcommand{\theHsection}{appendix.\Alph{section}}
\section*{Overview}
This supplementary document provides additional quantitative results and implementation details to support the main paper "Does Peer Observation Help? Vision-Sharing Collaboration for Vision-Language Navigation". The content is organized as follows:

\begin{itemize}
    \item \textbf{Appendix~\ref{A}: Training Details of the Spatial Overlap Detector.} The pair construction procedure, node feature representation, network architecture, and optimization settings used to train the spatial overlap detector.
    \item \textbf{Appendix~\ref{B}: Efficiency Analysis.} Parameter and runtime overhead introduced by our framework on top of the base navigator.
    \item \textbf{Appendix~\ref{C}: Prior-Based Episode Pairing.} Detailed algorithmic procedure for grouping and pairing episodes to maximize spatial overlap.
    \item \textbf{Appendix~\ref{D}: Prompt for MapGPT with Vision-Sharing.} The complete system prompt and specific modifications introduced to incorporate peer observation context.
\end{itemize}

\section{Training Details of the Spatial Overlap Detector}
\label{A}

We provide here the full training recipe of the spatial overlap detector introduced in the main paper.

\textbf{Pair construction.} Training pairs are generated offline from the Matterport3D~\cite{Matterport3D} connectivity graphs, from which we read the ground-truth 3D position of every viewpoint. A pair of viewpoints from the same scene is labeled positive when their Euclidean distance is below the overlap radius $R$, following Eq.~1 in the main paper. Negative pairs consist of same-scene pairs whose distance exceeds $R$ by a margin of at least $5\,\text{m}$, together with cross-scene pairs, which account for roughly $30\%$ of the negatives. In total we mine $1.2$ million training pairs from the R2R~\cite{anderson2018vision} train scenes and $0.2$ million evaluation pairs from the val unseen scenes, so the two sets share no scene and the reported detection metrics are free of leakage.

\textbf{Node features.} Each viewpoint is represented by the mean of its ViT view features over a random subset of $12$ to $30$ of the $36$ discretized views. This random subsetting simulates the partial observability at deployment time, where a node embedding in the topological map averages only the views the agent has actually observed, so the training-time representation matches the one the detector receives during navigation.

\begingroup
\emergencystretch=2em
\textbf{Architecture and optimization.} The detector projects the two $768$-dimensional node features into $512$-dimensional tokens, processes the resulting two-token sequence with a $3$-layer Transformer encoder with $8$ attention heads, and predicts the overlap logit from the concatenated outputs through a two-layer MLP head. We train with a class-balanced binary cross entropy loss, where the positive class weight is set to the negative-to-positive ratio of the training set. Optimization uses AdamW with a learning rate of $4\times10^{-4}$, weight decay of $10^{-4}$, a cosine learning rate schedule, and a batch size of $2048$ for up to $15$ epochs with early stopping.
\par\endgroup

\section{Efficiency Analysis}
\label{B}

We quantify the overhead our framework introduces on top of the base navigator, taking DUET~\cite{chen2022think} in the two-agent setting on the R2R val unseen split as the reference configuration.

\textbf{Parameter overhead.} The spatial overlap detector is the only additional module and adds $6.83$M parameters, which corresponds to $3.78\%$ of DUET's $180$M parameters. It is invoked for overlap detection between topological maps.

\textbf{Runtime overhead.} Measured on a single NVIDIA RTX A6000 with a batch size of $8$, per-episode inference time increases from $17.6$\,ms to $29.4$\,ms ($+11.8$\,ms), which covers detector scoring over candidate node pairs and the subsequent graph fusion. The geometry verification step operates on geodesic distances that are already maintained within each agent's own topological map, so its additional cost is negligible. The overall overhead is negligible relative to real-world robot action execution.

\section{Prior-Based Episode Pairing}
\label{C}
\begin{algorithm}[h]
\caption{Prior-Based Episode Pairing}
\label{alg:pairing}
\begin{algorithmic}[1]
\REQUIRE Episode set $\mathcal{D} = \{d_1, d_2, \dots, d_K\}$, random seed $s$
\ENSURE Paired lists $\mathcal{L}_1$, $\mathcal{L}_2$
\STATE Group $\mathcal{D}$ by scan into $\{\mathcal{D}_{e}\}_{e \in \mathcal{E}}$
\FOR{each scan group $\mathcal{D}_{e}$}
    \STATE Shuffle $\mathcal{D}_{e}$ with seed $s$
    \WHILE{$\mathcal{D}_{e} \neq \emptyset$}
        \STATE Pop first episode $d_a$ from $\mathcal{D}_{e}$
        \IF{$\mathcal{D}_{e} = \emptyset$}
            \STATE $\mathcal{L}_1 \leftarrow \mathcal{L}_1 \cup \{d_a\}$, \ $\mathcal{L}_2 \leftarrow \mathcal{L}_2 \cup \{d_a\}$, \ \textbf{break}
        \ENDIF
        \STATE $\mathcal{C} \leftarrow \{d_b \in \mathcal{D}_{e} \mid \mathrm{path}(d_a) \neq \mathrm{path}(d_b) \wedge \mathrm{start}(d_a) \neq \mathrm{start}(d_b)\}$
        \IF{$\mathcal{C} \neq \emptyset$}
            \STATE $d_b^* \leftarrow \arg\max_{d_b \in \mathcal{C}} |\mathrm{path}(d_a) \cap \mathrm{path}(d_b)|$
            \STATE Remove $d_b^*$ from $\mathcal{D}_{e}$
            \STATE $\mathcal{L}_1 \leftarrow \mathcal{L}_1 \cup \{d_a\}$, \ $\mathcal{L}_2 \leftarrow \mathcal{L}_2 \cup \{d_b^*\}$
        \ELSE
            \STATE $\mathcal{L}_1 \leftarrow \mathcal{L}_1 \cup \{d_a\}$, \ $\mathcal{L}_2 \leftarrow \mathcal{L}_2 \cup \{d_a\}$
        \ENDIF
    \ENDWHILE
\ENDFOR
\RETURN $\mathcal{L}_1$, $\mathcal{L}_2$
\end{algorithmic}
\end{algorithm}

Algorithm~\ref{alg:pairing} details the prior-based episode pairing procedure used in our experiments. Given the full episode set $\mathcal{D}$, we first group episodes by their environment scan. Within each scan group, we greedily pair episodes that (1) correspond to different trajectories with different starting viewpoints, and (2) maximize the number of overlapping viewpoints between their ground-truth paths. This encourages spatial proximity between paired agents, increasing the likelihood of spatial overlap during navigation. When a scan group contains an odd number of episodes or no valid partner can be found, the remaining episode is paired with itself, effectively reducing to the original baseline.

\section{Prompt for MapGPT with Vision-Sharing}
\label{D}

We retain the original MapGPT system prompt and introduce minimal modifications to incorporate peer observation context. The complete system prompt is shown below. Text inside green boxes indicates our additions. The modifications are limited to three aspects: (1) informing the agent that images prefixed with \texttt{other\_*} come from peer agents, (2) introducing a \texttt{Supplementary\_Map} field containing the topological structure shared by peer agents, and (3) including \texttt{Supplementary\_Map} in the reasoning and planning instructions.

\definecolor{promptbg}{RGB}{245,247,250}
\definecolor{promptframe}{RGB}{180,190,210}
\definecolor{addedbg}{RGB}{210,240,210}
\definecolor{addedframe}{RGB}{100,180,100}
\definecolor{sectioncolor}{RGB}{50,100,180}

\newtcolorbox{promptbox}[1][]{
  enhanced,
  breakable,
  colback=promptbg,
  colframe=promptframe,
  fonttitle=\bfseries\sffamily,
  title=#1,
  coltitle=white,
  colbacktitle=sectioncolor,
  boxrule=0.8pt,
  arc=3pt,
  left=8pt, right=8pt, top=6pt, bottom=6pt,
  fontupper=\small,
}

\newtcolorbox{addedbox}{
  enhanced,
  colback=addedbg,
  colframe=addedframe,
  boxrule=0.5pt,
  arc=2pt,
  left=6pt, right=6pt, top=3pt, bottom=3pt,
  fontupper=\small,
}

\newcommand{\sectionlabel}[1]{{\sffamily\bfseries\color{sectioncolor}#1}}
\newcommand{\added}[1]{\colorbox{addedbg}{#1}}

\begin{promptbox}[System Prompt for MapGPT with Vision-Sharing]

\sectionlabel{[Task Background]} \\[0.3em]
You are an embodied robot that navigates in the real world. You need to explore between some places marked with IDs and ultimately find the destination to stop. At each step, a series of images corresponding to the places you have explored and have observed will be provided to you.

\begin{addedbox}
Images prefixed with `Image other\_*' denote locations derived from the other agents' trajectories or observations. (note that these may not always be present)
\end{addedbox}

\sectionlabel{[Input Definitions]} \\[0.3em]
\textbf{Instruction} is a global, step-by-step detailed guidance, but you might have already executed some of the commands. You need to carefully discern the commands that have not been executed yet.

\textbf{History} represents the places you have explored in previous steps along with their corresponding images. It may include the correct landmarks mentioned in the Instruction as well as some past erroneous explorations.

\textbf{Trajectory} represents the ID info of the places you have explored. You start navigating from Place 0.

\textbf{Map} refers to the connectivity between the places you have explored and other places you have observed.

\begin{addedbox}
\textbf{Supplementary\_Map} refers to the connectivity between the places you have not explored, which shared intelligence from other agents. (note that Supplementary\_Map may not always be present). It includes unexplored areas, effectively expanding your situational awareness. Use this information to refine your navigation strategy and explore beyond your direct line of sight. Places prefixed with `other\_*' denote locations derived from the other agents' trajectories or observations.
\end{addedbox}

\textbf{Supplementary Info} records some places and their corresponding images you have ever seen but have not yet visited. These places are only considered when there is a navigation error, and you decide to backtrack for further exploration.

\textbf{Previous Planning} records previous long-term multi-step planning info that you can refer to now.

\textbf{Action Options} are some actions that you can take at this step.

\vspace{0.5em}
\sectionlabel{[Output Requirements]} \\[0.3em]
For each provided image of the places, you should combine the Instruction and carefully examine the relevant information, such as scene descriptions, landmarks, and objects. You need to align Instruction with History (including corresponding images) to estimate your instruction execution progress and refer to Map for path planning. Check the Place IDs in the History and Trajectory, avoiding repeated exploration that leads to getting stuck in a loop, unless it is necessary to backtrack to a specific place. If you can already see the destination, estimate the distance between you and it. If the distance is far, continue moving and try to stop within 1 meter of the destination.

Your answer should be JSON format and must include three fields: Thought, New Planning, and Action. You need to combine Instruction, Trajectory, Map, \added{Supplementary\_Map,} Supplementary Info, your past History, Previous Planning, Action Options, and the provided images to think about what to do next and why, and complete your thinking into Thought.

Based on your Map, \added{Supplementary\_Map,} Previous Planning and current Thought, you also need to update your new multi-step path planning to New Planning.

At the end of your output, you must provide a single capital letter in the Action Options that corresponds to the action you have decided to take, and place only the letter into Action, such as ``Action'': ``A''.
\end{promptbox}

{\small\sffamily \textbf{Legend:} \colorbox{addedbg}{Green boxes} = modifications introduced for vision-sharing. All other content is identical to the original MapGPT prompt.}

\end{document}